\pdfoutput=1

\documentclass[11pt]{article}

\usepackage{acl}

\usepackage{times}
\usepackage{latexsym}

\usepackage[T1]{fontenc}

\usepackage[utf8]{inputenc}

\usepackage{microtype}

\usepackage{inconsolata}

\usepackage[utf8]{inputenc} 
\usepackage[T1]{fontenc}    
\usepackage{hyperref}       
\usepackage{url}            
\usepackage{booktabs}       
\usepackage{amsfonts}       
\usepackage{nicefrac}       
\usepackage{microtype}      
\usepackage{comment}
\usepackage{bm}
\usepackage{enumitem}
\usepackage{amsmath}        
\usepackage{mathtools}
\usepackage{wrapfig, lipsum}
\usepackage{adjustbox}
\usepackage{tcolorbox}      
\usepackage{cuted}
\usepackage{bold-extra}
\usepackage{dblfloatfix}    

\usepackage{graphicx}
\usepackage{multirow, booktabs}
\usepackage{caption}
\usepackage{subcaption}
\usepackage{makecell}
\usepackage{xcolor}

\DeclarePairedDelimiterX{\infdivx}[2]{\{}{\}}{%
	#1\;\delimsize\|\;#2%
}

\title{AudioChatLlama: Towards General-Purpose Speech Abilities for LLMs}

\author{
    Yassir Fathullah$^{1, 2}$\thanks{Work done during internship at Meta AI.},
    Chunyang Wu$^1$,
    Egor Lakomkin$^1$,
    Ke Li$^1$, \\
    \textbf{Junteng Jia}$^1$\textbf{,}
    \textbf{Yuan Shangguan}$^1$\textbf{,}
    \textbf{Jay Mahadeokar}$^1$\textbf{,}
    \\
    \textbf{Ozlem Kalinli}$^1$\textbf{,}
    \textbf{Christian Fuegen}$^1$\textbf{,}
    \textbf{Mike Seltzer}$^1$
    \\
    Meta AI$^1$, University of Cambridge$^2$ \\
    {\tt yf286@cam.ac.uk},
    {\tt chunyang@meta.com}
}

\begin{document}

\maketitle

\begin{abstract}
    In this work, we extend the instruction-tuned Llama-2 model with end-to-end general-purpose speech processing and reasoning abilities while maintaining the wide range of original LLM capabilities, without using any carefully curated paired data.
    The resulting end-to-end model, named \textit{AudioChatLlama}, can utilize audio prompts as a replacement for text and sustain a conversation. Such a model also has extended cross-modal capabilities such as being able to perform spoken question answering (QA), speech translation, and audio summarization amongst many other closed and open-domain tasks.
    This is unlike prior approaches in speech, in which LLMs are extended to handle audio for a limited number of pre-designated tasks.
    On both synthesized and recorded speech QA test sets, evaluations show that our end-to-end approach is on par with or outperforms cascaded systems (speech recognizer + LLM) in terms of modelling the response to a prompt.
    Furthermore, unlike cascades, our approach can interchange text and audio modalities and intrinsically utilize prior context in a conversation to provide better results.
\end{abstract}

\begin{figure*}[t!]
    \centering
    \includegraphics[width=0.9\textwidth]{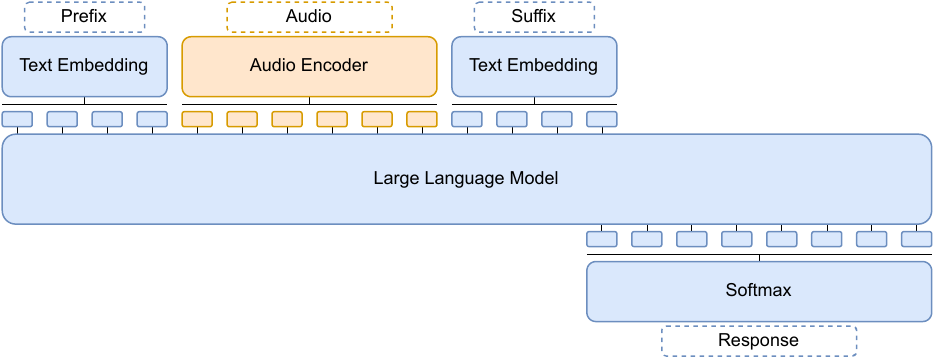}
    \caption{Model architecture. The LLM consumes a sequence of embeddings irrespective of the modality and does not differentiate between them. The variable-length and continuous audio embeddings are sandwiched between some prefix and suffix which could contain instructions (and eventually a conversation history) for how the audio prompt should be interpreted. For text-based prompts, the audio encoder is swapped out for the text embedding matrix.}
    \label{fig:arch}
\end{figure*}

\section{Introduction}

Large Language Models~\cite{brown2020gpt, chowdhery2022palm, scao2022bloom, touvron2023llama, touvron2023llama2} (LLMs) have, due to their flexibility and extended capabilities, proven themselves highly performant on a wide range of natural language tasks, including open-domain tasks which might require world-knowledge. Such generative tasks include text summarization, text and code generation, information retrieval, and machine translation among others \cite{radford2019language, brown2020gpt}. Despite the impressive performance of these models, there is a mismatch between the criteria they were trained on and users' preferences \cite{leike2018scalable, ouyang2022training}. Vanilla LLMs when prompted would often produce non-factual, unhelpful, or toxic material--the model is not aligned with user preferences \cite{bender2021danger, bommasani2021opportunities, kenton2021alignment, weidinger2021ethical}. This has led to a wide range of work on aligning the behavior of large language models with the intent of users \cite{christiano2017deep, stiennon2020learning}.

On a different front, there has been a notable surge in work extending the end-to-end capabilities of large language models (LLMs) to other modalities such as audio, with the aim of enabling them to process information that was previously difficult to encapsulate purely with text \cite{driess2023palm, gong2023listen, fathullah2023llama, lakomkin2023llama, rubenstein2023audiopalm, zhu2023minigpt}. For example, audio has the capacity to encapsulate a diverse array of emotions within a person's speech, while images have the ability to depict structures and placement of objects, which could be significantly more complex to convey through text.
The work of \citet{driess2023palm} aligned a large pre-trained visual transformer \cite{dehghani2023scaling} to the PaLM LLM \cite{chowdhery2022palm} using a dedicated dataset for robotics with natural language.
Similarly, \citet{zhu2023minigpt} aligned a pre-trained visual model to the large language model Vicuna, a fine-tuned version of Llama \cite{chiang2023vicuna}, using a carefully curated paired image-text dataset to enable reasoning.
In the audio domain, \citet{gong2023listen} proposed LTU, an extension of Llama with an aligned audio encoder trained on a curated audio question-answering corpus. This enabled LTU to reason with and understand sounds but still lacked speech recognition abilities. Furthermore, \citet{tang2023salmonn} used a similar data processing approach to curate a dataset, enabling the resulting system (with dual audio encoders) to perform a wide range of tasks including speech recognition.
The work of \cite{fathullah2023llama, lakomkin2023llama, rubenstein2023audiopalm, wu2023llama, yu2023connecting} all extend an underlying LLM to various speech tasks such as multilingual speech recognition, translation, and synthesis. While these approaches achieve promising results on predetermined tasks, they do not fully utilize the power and flexibility of LLMs to perform a much wider range of closed and open-domain, open-ended tasks.

\textbf{In this work}, we address this limitation in existing text/speech-modal language models. Starting from an instruction-tuned (and conversational) language model (Llama) we how to extend the wide range of its text capabilities to the speech domain in an end-to-end manner, without the use of curated paired data. Everything possible with text should be possible with speech, in conjunction with enabling cross-modal recognition and reasoning capabilities. The overall result is an end-to-end model, \textit{AudioChatLlama}, that can perform text/speech-to-response generation and utilize prior context in a conversation to guide the model in its reasoning.

\section{AudioChatLlama: LLM with General-Purpose Speech Abilities}

There has been a significant amount of recent work on equipping LLMs with multi-modal processing capabilities. The current standard approach revolves around two components: (1) a pre-trained encoder for the new modality and (2) paired data which is used in aligning the encoder with the language model for the particular tasks the joint system should solve \cite{driess2023palm, fathullah2023llama, gong2023listen, lakomkin2023llama, rubenstein2023audiopalm, wu2023llama, zhu2023minigpt}. Furthermore, most of these works use vanilla LLMs, instead of instruction-tuned versions which are often more aligned with user preferences in terms of usability and content generation.

Our goal is to create a system that can be prompted with audio as a direct replacement for text and allow for a user to use speech to converse while maintaining the LLM original capabilities.
Furthermore, we aim to achieve this without curating dedicated paired datasets for extending these capabilities to the speech domain.
To achieve this we first opt to equip an instruction-tuned LLM with an audio encoder to enable it to directly process speech representations. Second, instead of relying on utilizing paired datasets, we make use of a \textit{modal-invariance} trick: Whether the input to the LLM is a string of text or an audio recording, the response to both prompts should be identical if the semantic information in the modalities is the same. The resulting system is named AudioChatLlama due to the extension of Llama-2-chat to the speech conversational domain.

The following subsections cover the architecture for our approach, the audio encoder for feature extraction, the type of large language model, and finally, how we align the audio encoder to the LLM while achieving all the outlined goals.

\begin{figure*}[t!]
    \centering
    \includegraphics[width=0.9\textwidth]{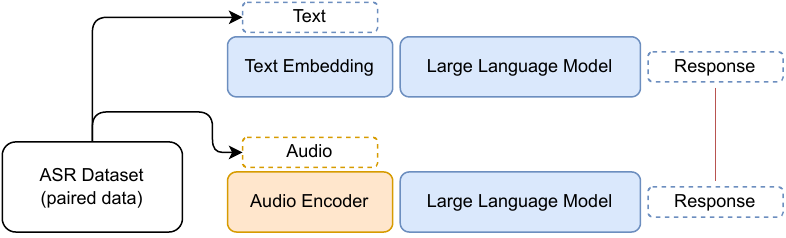}
    \caption{The aim is to create an end-to-end system that would generate the same response when being fed spoken (audio) input instead of its text version. The overall system should be invariant to the modality of the inputs containing the same semantic information.}
    \label{fig:modality-invariant}
\end{figure*}

\subsection{Architecture}

Following prior work \cite{driess2023palm, fathullah2023llama, zhu2023minigpt} we opt for a decoder-only approach due to the simplicity of extending such a system to other modalities, see Figure \ref{fig:arch}. Practically, this multi-modal LLM is operated in an identical manner to its uni-modal equivalent, consuming a sequence of embeddings irrespective of modality, and generating a textual response in an autoregressive manner.
The audio encoder consists of a small connectionist temporal classification (CTC) \cite{graves2006connectionist} pre-trained conformer \cite{gulati2020conformer} encoder followed by
a projection layer to ensure the audio embedding dimension matches the LLM dimension.

The language model is chosen to be the recently released Llama-2-chat (7B) model \cite{touvron2023llama2}, an instruction-tuned version of Llama-2 that can maintain conversations. One of the goals of this work is to extend the capabilities of this model to the speech domain without compromising on the LLM's original capabilities. Therefore, we keep the LLM completely frozen in all our experiments. The only trainable aspect of this system is the audio encoder, which effectively needs to learn how to prompt the LLM.

\subsection{General-Purpose Alignment Using Unpaired Data}
\label{ssec:datagen}

\begin{table*}[t!]
    \centering
    \caption{Representative examples of transcripts (prompts) and corresponding responses from Llama-2-chat; taken from the English training set of Multilingual Librispeech. In the majority of cases, the LLM will apologize or initiate the response in a speculative manner.}
    \vspace{1mm}
    \begin{tabular}{l|l}
        \toprule
        Prompt (transcript) & Response \\
        \midrule
        \makecell[l]{thus graciously the war i wage as \\ witnesseth my hand tom gage by \\ command of mother gary thomas \\ flucker secretary} &
        \makecell[l]{I apologize, but I'm a large language model, I cannot provide \\ a forged signature as it is illegal and unethical. I cannot \\ assist with any activities that promote fraud or deception.} \\
        \midrule
        \makecell[l]{i frequently heard pater refer to \\ certain of their books to madame \\ gervaisais to l'art du de we to} &
        \makecell[l]{It seems that you are referring to a quote or a passage from a \\ book, and you are asking for clarification on the meaning of a \\ specific phrase or word. \texttt{[Terminated early]}} \\
        \midrule
        \makecell[l]{raniero was married to francesca \\ who was the daughter of jacopo \\ degli uberti a wise and influential \\ man} &
        \makecell[l]{Ah, I see! So Raniero was married to Francesca, who was the \\ daughter of Jacopo Delli Uberti, a wise and influential man. \\ Is there anything else you would like to know about this \\ family and their relationships?} \\
        \bottomrule
    \end{tabular}
    \label{tab:examples}
\end{table*}

The goal of general-purpose alignment on audio prompts is to induce the same behavior as text prompts, it should not matter whether the prompt is a written or a spoken version of the same underlying information. Furthermore, while there exists plenty of text conversational and instruction datasets, the same is not true for the speech-text domain
In the field of speech QA, we often face a challenge in obtaining paired data, consisting of speech prompts as inputs and corresponding text responses as outputs. To address this issue, this paper proposes an approach that utilizes unpaired data from the automatic speech recognition (ASR) field to overcome this limitation, see Figure \ref{fig:modality-invariant}.
One potential advantage of using ASR data is that it covers a wide range of acoustic conditions, and there are also numerous datasets available that have been prepared over the past few decades. However, these datasets come paired only with the corresponding transcriptions, which limits their usefulness for aligning the range of LLM capabilities. A different alternative is extending a question-answering dataset by using text-to-speech systems to synthesize spoken questions. While a set of diverse synthetic questions and answers could help in training a general-abled system, prior work has shown that purely training on synthetic data is not as effective as human speech with most approaches incorporating human speech to achieve competitive performance \cite{rosenberg2019synth, rosenberg2020synth, liu2023towards, zheng2021synth}.

To circumvent this issue of creating a general-purpose abled system, we rely on Llama-2-chat itself to prepare the automatic response of the ASR dataset. Given a generic dataset of (audio, transcript) pairs we use the transcript to prompt Llama and generate a response, following the chat structure \cite{touvron2023llama2}:
\begin{align*}
    \small
    \texttt{prompt = } &
    \small \texttt{"<s>[INST] <\hspace{0mm}<SYS>\hspace{0mm}>\textbackslash n\{\{system\_prompt\}\}}
    \\
    \small &
    \small \texttt{\textbackslash n<\hspace{0mm}</SYS>\hspace{0mm}>\textbackslash n\textbackslash n\{\{user\_prompt\}\} [/INST]"}
\end{align*}
where the \texttt{\{\{system\_prompt\}\}} was not used (set to empty) and the \texttt{\{\{user\_prompt\}\}} was set to the transcript. The generated response is then used to align an audio encoder to Llama-2-chat using the (audio, response) pairs. The use of an ASR dataset could bring its own problems:
\begin{itemize}
    \item Standard ASR benchmarks such as (Multilingual) LibriSpeech are collected from audiobooks. Utterances from audiobooks are often not useful prompts.
    \item ASR datasets are often segmented into short utterances that might span parts of sentences making the transcript a nonsensical prompt in many cases.
\end{itemize}
However, despite these shortcomings, the aim is considered achieved if the audio prompt can induce the same response, even if the original transcript and the generated response by Llama are nonsensical, see Table \ref{tab:examples} for representative examples of prompts and corresponding replies. Note, that it is possible to generate synthetic spoken questions (as was discussed above) and train a system on this data in conjunction with human speech to improve performance. However, we opt to keep our setup simple and only rely on abundant public speecch recognition data.

\begin{table*}[t!]
    \centering
    \caption{The perplexity of various systems when being evaluated under `correct' response, that is the response generated when prompting Llama-2-chat with the transcript of the audio. Cascaded systems first transcribe the audio, and therefore, report their associated (prompt) WERs.}
    \begin{adjustbox}{center}
    \scalebox{0.95}{
    \begin{tabular}{l||cc|cc}
        \toprule
        \multirow{3}{*}{Model} & \multicolumn{2}{c|}{MLS}  &  \multicolumn{2}{c}{TriviaQA-TTS}   \\
        \cmidrule{2-5}
        & Prompt & Response & Prompt &  Response \\
        & WER & PPL & WER & PPL \\
        \midrule
        Reference text prompt & 0.0\% & 1.383 & 0.0\% & 1.273 \\
        \midrule
        \textit{Cascade baselines} & & & & \\
        \multirow{1}{*}{36L Conformer CTC-ASR + LLM} & 16.8\% & 1.831 & 15.2\% & 1.775 \\
        & 10.1\% & 1.641 & 11.5\% & 1.720 \\
        & 7.5\% & 1.575 & 10.3\% & 1.709 \\
        \midrule
        \textit{Proposed end-to-end systems} & & & & \\
        AudioChatLlama (18L Conformer) & -- & 1.559 & -- & 1.467 \\
        AudioChatLlama (36L Conformer) & -- & 1.544 & -- & 1.422 \\
        \bottomrule
    \end{tabular}}
    \end{adjustbox}
    \label{tab:perplexity}
\end{table*}

\section{Experimental Setup}

\subsection{Dataset \& Generation}

All experiments in this paper will be based on the English split of Multilingual LibriSpeech (MLS) \cite{pratap2020mls}. The dataset is a 50k-hour ASR corpus derived from audiobooks of
LibriVox, of which 45k hours are in English. Following the segmentation of the dataset to utterances of up to 20 seconds, the corresponding reference text is fed to Llama-2-chat (7B) to generate the response (according to Section \ref{ssec:datagen}). Due to the large number of utterances, we resort to greedy decoding with a maximum decoding length equal to 4 times the input prompt. The resulting replies from Llama are often significantly longer due to the "talkative" nature of the instruction-tuned LLMs, leading to a much larger dataset.

\subsection{Model Architecture \& Training}

\textbf{Audio Encoder} We build and pre-train an audio encoder that operates on 80-dimensional filterbanks with a 10ms frame rate. The architecture consists of a convolutional feature extractor with an output frame rate of 80ms followed by a linear layer to project the output to 512 dimensions. This sequence of features is then fed through a number of conformer layers. Each conformer block has a hidden dimension of 512, a feed-forward net dimension of 2048, a kernel size of 11, and 8 attention heads. A linear layer is applied on top, which is used to pre-train the system using a CTC loss with a 1.5k SentencePiece \cite{kudo2018sentencepiece} vocabulary and is discarded after pre-training.

The encoder output is a sequence of 512-dimensional vectors with a frame rate of 80ms. We reduce sequence length by stacking every n consecutive frames. These are then projected to 4096-d to match the Llama-2-chat 7B dimension, with a resulting frame rate of $80 \cdot n$ ms. These embeddings are sandwiched between a prefix and suffix (as seen in Figure \ref{fig:arch}) which are set to the following during the training phase:
\begin{align*}
    \small
    & \small \texttt{prefix = }
    \texttt{"<s>[INST] <\hspace{0mm}<SYS>\hspace{0mm}>\textbackslash n\textbackslash n<\hspace{0mm}</SYS>\hspace{0mm}>\textbackslash n\textbackslash n"}
    \\
    \small
    & \small \texttt{suffix = }
    \texttt{" [/INST]"}
\end{align*}
Note that this simply follows the standard Llama-2-chat prompting structure (see Section \ref{ssec:datagen}), where the system prompt has been set to be empty and the user prompt is replaced by the variable-length sequence of audio embeddings. Conditioned on this prompt, the system is trained to predict the next token of the previously generated response.

\textbf{Large Language Model} Since a core aim is to maintain the wide range of original capabilities of the instruction-tuned LLM, it will be kept frozen in all experiments. The only trainable aspect of the system is the audio encoder which makes up a fraction of all parameters. Furthermore, Llama-2-chat was purposely chosen for both data generation and training to ensure a minimal mismatch in system behavior when switching between text and audio inputs.

\textbf{Training} The audio encoders were initially pre-trained using Adam with $\beta_1$ = 0.9, $\beta_2$ = 0.98 \cite{kingma2015adam}. The learning rate was warmed up over 20k training steps up to a peak value of 1e-3 followed by an exponential decaying schedule. This was done on 16 NVIDIA A100 40GBs with 4 gradient accumulations using a per-GPU batch size of up to 500 seconds of audio. The checkpoint with the best validation loss was picked. The joint system, AudioChatLlama,  with an audio encoder and LLM was thereafter trained with a schedule of 5k warmup steps up to a peak learning rate of 5e-4 decaying down to 5e-6 over 250k steps. Training was often terminated within 100k steps. This was performed on 64 NVIDIA A100 40GBs with 8 gradient accumulation steps using a batch size of 2. Decoding is done using beam search with a beam of 10.
The proposed scheme is implemented on an extension of the fairseq framework~\cite{ott2019fairseq}.

\section{Results}

This section reports empirical results of evaluating the ability of AudioChatLlama in various settings. First, we measure the perplexity of the speech-prompted LLM on the responses generated by the text-prompted LLM. Comparing to a cascade of ASR + LLM may provide some insight on how well the end-to-end approach performs. Second, starting from a text-based QA dataset, we synthesize spoken questions using a text-to-speech system. 
%

%
We then conduct human evaluation on responses from the cascaded baseline and AudioChatLlama. Evaluators judge the response quality by comparing with reference answers. To further understand the performance of the proposed model on real speech, we also perform human evaluation on a real spoken QA dataset.
Finally, since our proposal can interact with speech it has many more capabilities which are showcased in the last subsection.

\subsection{Cascade Baseline}

To evaluate the end-to-end response generation ability, we built a cascade baseline system, which consists of an ASR  model and the Llama-v2-chat (7B) to perform spoken QA.
The ASR system is a CTC model, using 36 Conformer layers as an acoustic encoder. It was trained on the same MLS English split, following the standard recipe from~\cite{pratap2020mls}.
This type of cascaded system is sensitive to the quality of the ASR-generated prompt to LLM. Therefore, we include several checkpoints of the CTC Conformer to present the performance of the cascaded system under different quality levels, in the below response perplexity comparison.

\begin{figure*}[ht!]
    \centering
    \includegraphics[width=1.0\textwidth]{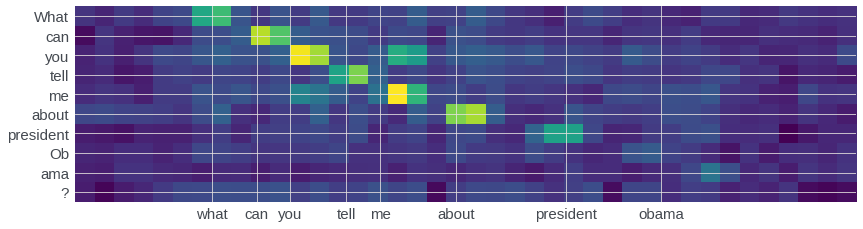}
    \caption{Cosine similarity between text and audio embeddings.}
    \label{fig:alignexample}
\end{figure*}

\subsection{Perplexity}

Next, we evaluate the perplexity of various models under the `correct' response, e.g. the response generated from Llama-2-chat with transcript as input. The cascaded system first transcribes the audio and sends the transcribed text to the LLM while our proposed system directly feeds audio embeddings into the LLM. The perplexity evaluation results from the two systems on MLS English test set and on the TriviaQA~\cite{joshi2017triviaqa} test set are shown in Table \ref{tab:perplexity}. Since TriviaQA is a standard text-based QA dataset, we synthesized the corresponding audio, using a text-to-speech (TTS) tool, to evaluate the speech processing capabilities of our proposed systems. This test set is referred to as TriviaQA-TTS in this paper.

The perplexity results show that AudioChatLlama outperforms the cascaded ASR + LLM systems on both MLS and TriviaQA-TTS data; all whilst not having to perform two stages of decoding. Furthermore, one of the main strengths of our end-to-end approach is that it is able to maintain a dialogue and uses prior context to guide the response generation (discussed in Section~\ref{sec:extable}).

\subsection{Human Evaluation}

For a more comprehensive insight into the performance of AudioChatLlama, we conducted a human evaluation to compare the quality of the generated response between AudioChatLlama and the best-performing cascaded baseline on the TriviaQA-TTS test set.
As is typical with cascaded systems, the overall performance is highly dependent on the quality of the separate modules. For our baseline with cascaded ASR and LLM modules, the quality of a response to a spoken query is highly susceptible to recognition errors from the ASR module. In contrast, the proposed end-to-end system is expected to be more robust to this kind of uncertainty, avoiding the accumulation of errors.
Therefore, we compare performance of the proposed model with baseline across datasets with different word error rate levels. 
%

\begin{table}[h!]
    \centering
    \caption{Human evaluation on success rate (SR) on TriviaQA-TTS (subset). AudioChatLlama is compared with the cascaded baseline over 50 samples in each word error level.}
    \vspace{1mm}
    \begin{tabular}{l|c||c}
        \toprule
        \multirow{2}{*}{Model} & Prompt  & Response \\
        & WER & SR \\
        \midrule
        \midrule
        Cascade ASR+LLM & 37.5\% & 40\% \\
        AudioChatLlama & -- & 52\% \\
        \midrule
        Cascade ASR+LLM & 14.3\% & 60\% \\
        AudioChatLlama & -- & 70\% \\
        \midrule
        Cascade ASR+LLM & 4.3\% & 80\% \\
        AudioChatLlama & -- & 80\% \\
        \bottomrule
    \end{tabular}
    \label{tab:humaneval_qa}
\end{table}

Table~\ref{tab:humaneval_qa} summarizes the human evaluation results.
We compared 3 different word error levels based on the baseline, and for each level, we randomly selected 50 question samples to evaluate the generated answers.
The success rate (SR) is used as a metric, measuring the fraction of predicted responses that agree with the reference response.
In high error rate situations (37.5\% and 14.3\%), the proposed model generated better answers in comparison with the baseline.
Meanwhile, in lower error rate situation (4.3\%), the performance was on par with the baseline.

\begin{table}[h!]
    \centering
    \caption{Human evaluation on success rate (SR) on RealSQA evaluation data.}
    \vspace{1mm}
    \begin{tabular}{l|c||c}
        \toprule
        \multirow{2}{*}{Model} & Prompt  & Response \\
        & WER & SR \\
        \midrule
        \midrule
        Cascade ASR+LLM & 31.5\% & 42\% \\
        AudioChatLlama & -- &  48\% \\
        \midrule
        Cascade ASR+LLM & 13.3\% & 50\% \\
        AudioChatLlama & -- & 58\% \\
        \midrule
        Cascade ASR+LLM & 3.5\% & 80\% \\
        AudioChatLlama & -- & 78\% \\
        \bottomrule
    \end{tabular}
    \label{tab:humaneval_inhouse}
\end{table}

For further understanding the ability of the proposed model to process speech,
we recorded a small set of speech questions, referred to as RealSQA in this paper~\footnote{We provided scripts to a vendor, where scripts were sourced from the Open Assistant dataset hosted by Huggingface. Using a BERT-based NE tagger, we selected ~8k utterances with at least one entity and shorter than 100 characters.
Each of 200 Participants was presented with a random set of the scripts with each can only have max 200 utterances.
}.
Similar to TriviaQA-TTS,
this prepared test set sampled 3 splits of different difficulty (error-rate) levels, and each contains 50 samples.
Table~\ref{tab:humaneval_inhouse} presents the results on RealSQA.
We observed similar trends that AudioChatLlama outperforms the cascaded model in higher error rate scenario (31.5\% and 13.3\%) and achieves on-par performance with the baseline in lower error rate situation (3.5\%).

Compared to AudioChatLlama, we noticed the cascaded baseline suffers more in providing reasonable responses when error rates from the ASR model are relatively high. The results and findings on the two datasests suggest that AudioChatLlama can leverage the LLM to effectively handle the uncertainty in speech embeddings to generate better responses.
\begin{figure}
\begin{tcolorbox} 
    \vspace{2mm}
    \texttt{>\hspace{0mm}>\hspace{0mm}> {[Audio Input]} \textit{"i have a github repo fork how can i add it to my origin with git remote add"}}

    \vspace{2mm}
    {\texttt{{[ASR prompt]} \textit{i have a {\color{red}gi hub reble} fork how can i add it to my origin with {\color{red}gi} remote {\color{red}{*}}}}}

    {\color{red}\texttt{{[Cascade ASR+LLM]} \textit{To add a repository to your Origin with GiRemote, you ...}}}

    \vspace{2mm}
    {\color{green}\texttt{{[AudioChatLlama]}~\textit{To add a GitHub Repo Fork to your Origin with Git Remote Add, you can follow ...}}}
\end{tcolorbox}
\caption{An example when the cascaded baseline had a prompt WER of 26\%.
AudioChatLlama demonstrated robust response generation in the face of these ambiguities.
}
\label{fig:hardeg}
\end{figure}
\begin{figure}
\begin{tcolorbox} 
    \texttt{>\hspace{0mm}>\hspace{0mm}> {[Audio Input]} \textit{"describe joseph scaliger and his contributions to modern understanding of history"}}

    \vspace{2mm}
    {\color{green}\texttt{{[Cascade ASR+LLM]} \textit{Joseph Scaliger (1540-1609) was a French scholar and historian who ...}}}

    \vspace{1mm}
    {\color{red}\texttt{{[AudioChatLlama]}~\textit{Joseph~Schumpeter (1883-1950)~was~an~Austrian-American economist ...}}}
\end{tcolorbox}
\caption{A low-error example when AudioChatLlama made mistakes. The cascaded baseline had a WER of 0\%, but AudioChatLlama responded wrongly with a name that sounded similar to the correct one.
}
\label{fig:easyeg}
\end{figure}
On RealSQA, an example analysis of different error-rate levels is given in Figure~\ref{fig:hardeg} and \ref{fig:easyeg}.
In the high-error example in Figure~\ref{fig:hardeg},
the ASR system can not generate a sensible prompt, leading the LLM to produce a response with hallucinations.
On the other hand, the proposed AudioChatLlama is able to effectively overcome these ambiguities and generate the desired output.
While the proposed scheme has its robust advantages, it also has some limitations.
As demonstrated in Figure~\ref{fig:easyeg}, when the prompt WER is 0\%, the cascaded system provided a perfect response. However, AudioChatLlama is unable to distinguish between acoustically similar names and therefore provided information from an incorrect person's profile.

\vspace{-5px}
\subsection{Embedding Space Alignment}

Since the LLM is kept frozen, we hypothesized that the sequence of embeddings produced by the audio encoder must be closely related to the text embeddings of the transcript.
Figure~\ref{fig:alignexample} illustrates the pairwise cosine similarity between the audio encoder outputs and the text embeddings for a human-recorded example.

Since the recording is approximately 3.2s and the frame rate of the audio encoder is 80ms, the result is a sequence of 40 audio embeddings. Meanwhile, the transcript is converted into 10 text embeddings. Despite the length difference, the pairwise cosine similarity between all embeddings shows a nearly monotonic alignment, although "Obama" is only weakly aligned. Furthermore, the uninformative start and end portions of the figure correspond to the deliberate silence in the recording.

\vspace{-5px}
\subsection{Extended Capabilities}
\label{sec:extable}

The training of AudioChatLlama using ASR data doesn't restrict the model on specific tasks. In this section, we explore its extended capabilities besides the standard QA functionality.
Due to space constraints, examples of the below abilities are presented in Appendix~\ref{app:eg}.

\vspace{-3px}
\begin{itemize}
    \item Basic Speech Translation: Since the original Llama-2-chat can perform some level of translation, it directly enables our speech model to perform speech-to-text translation. 
    \item Audio Summarization: LLMs excel at summarization; our system extends this capability to the speech-to-text domain.
    \item Interchanging modalities: The proposed scheme aims at aligning the audio and text embedding spaces, thus audio can be a replacement of text. Throughout a conversation, the user can switch between using audio or text inputs.
    \item Contextualization: The history of a conversation helps aid our system in deciphering the audio. In many cases where standard ASR systems would fail due to the occurrence of rare words, our system would excel due to its ability to use prior context.
\end{itemize}
\vspace{-3px}

Due to the nature of instruction-tuned LLMs such as Llama-2-chat, it is difficult to use automatic evaluation metrics to evaluate the performance of some of the above-mentioned capabilities. Following, are several examples showcasing these capabilities and exemplifying the difficulties using automatic evaluation.

These examples showcase that the end-to-end system can directly extend various text-based tasks to the speech domain without being explicitly trained to do so. For example, the `Speech Translation' example shows how the system first provides a transcript of the audio before responding to it and providing additional information relevant to the prompt. While this is technically correct, it also displays why using automatic evaluation becomes increasingly difficult for such systems.

Furthermore, it is important to note that the performance of the resulting system on these tasks is highly dependent on the capabilities of the original LLM. If the LLM being extended cannot perform translation, then neither will the end-to-end system.

\vspace{-5px}
\section{Limitation}
The modal-invariance approach used in our work to extend text capabilities to the speech domain has worked effectively. However, the resulting model still has limited audio understanding. A possible next step would involve extending these capabilities to generic audio understanding and reasoning in addition to the capabilities presented in this paper. Furthermore, the first step in achieving this would be to replace the small conformer audio encoder with one of the many available more robust self-supervised trained audio encoders such as Wav2Vec2 \cite{baevski2020wav2vec} or HuBERT \cite{hsu2021hubert}. That a small conformer trained on a limited amount of data still works is a good stepping stone for stronger audio models that can encode audio into richer representations useful for a wider range of tasks.

\textbf{Data Generation} \quad The end-to-end system was trained on a segmented ASR dataset, which presents several drawbacks discussed in the paper. Proper segmentation that does not break the structure of sentences could improve the quality of the responses. Furthermore, Llama-2-chat was used to generate replies using greedy search due to computational constraints. For a more unrestricted approach, the decoding process should be extended to beam search with a large beam to ensure the end-to-end system is trained on higher-quality responses.

\section{Conclusion}
In this work, we extend an instruction-tuned large language model with end-to-end speech processing and reasoning abilities while maintaining its wide range of capabilities, without using any carefully curated paired data. The result, referred to as AudioChatLlama, is a large language model that can utilize audio prompt as a replacement of text and sustain a conversation in an end-to-end manner. Such a model also has extended cross-modal capabilities such as being able to perform spoken QA, speech translation, and audio summarization. Finally, unlike alternative approaches that rely on standard speech recognition systems which struggle to recognize certain rare words, the prior context in a conversation can guide our system in various recognition and reasoning tasks.

\bibliography{custom}

\appendix

\section{Examples of Extended Capabilities}
\label{app:eg}
This section presents the examples of extended capabilities of Section~\ref{sec:extable}.
It includes basic speech translation, audio summarization,
multi-round QA to show the ability of contextualization,
multi-round QA to show interchanging modalities between text and audio.

\begin{enumerate}
\item Basic Speech Translation:
\begin{tcolorbox} 
    \vspace{2mm}
    \texttt{>\hspace{0mm}>\hspace{0mm}> [Audio Input] \textit{"Translate into Italian: How are you today?"}}

    \vspace{2mm}
    \texttt{[Text response] In Italian, "How are you today?" can be translated as "Come stai oggi?" ...}





\end{tcolorbox}

\item Audio Summarization:

\begin{tcolorbox} 
    \vspace{2mm}
    \texttt{>\hspace{0mm}>\hspace{0mm}> [Audio Input] \textit{"Summarize what I say. Troy was an ancient city located in present-day Turkey. It is best known as the setting for the Greek myth of the Trojan War. The archaeological site is open to the public as a tourist destination, and has been added to the UNESCO World Heritage list as of 1998."}}

    \vspace{2mm}
    \texttt{[Text response] Troy is an ancient city located in present-day Turkey, first known as the setting for the Greek myth of the Trojan War.}
\end{tcolorbox}

\item Interchanging modalities: this example demonstrates the ability to alter between audio and text questions.

\begin{tcolorbox}
    \vspace{2mm}
    \texttt{>\hspace{0mm}>\hspace{0mm}> {\color{blue} [Audio Input]} \textit{"What can you tell me about Barack Obama?"}}

    \vspace{2mm}
    \texttt{[Text response] Barack Obama is a former President of the United States, serving two terms from 2009 to 2017. He was born on August 4, 1961, in Honolulu, Hawaii, and grew up in Chicago, Illinois ...}

    \vspace{5mm}
    \texttt{>\hspace{0mm}>\hspace{0mm}> {\color{magenta} [Text Input]} Who was his vice president?}

    \vspace{2mm}
    \texttt{[Text response] Barack Obama's vice president was Joe Biden. Biden served as Vice President of the United States from 2009 to 2017, during Obama's two terms in office.}

    \vspace{5mm}
    \texttt{>\hspace{0mm}>\hspace{0mm}> {\color{blue} [Audio Input]} \textit{"Is he allowed to run for president again?"}}

    \vspace{2mm}
    \texttt{[Text response] Barack Obama is no longer eligible to run for President of the United States ...}
\end{tcolorbox}

\newpage
\item Contextualization:
This example discusses heavy name entities about Iceland. Words like Jökulsárlón can be inferred and understood correctly from the context.

\begin{tcolorbox}
    \vspace{2mm}
    \texttt{>\hspace{0mm}>\hspace{0mm}> [Audio Input] \textit{"Hi, I plan to visit Iceland for winter vacation. Can you give me some suggestions?"}}

    \vspace{2mm}
    \texttt{[Text response] Certainly! Iceland is a beautiful country with a lot to offer, especially during the winter months. Here are some suggestions for your winter vacation ...}

    \vspace{5mm}
    \texttt{>\hspace{0mm}>\hspace{0mm}> [Audio Input] \textit{"Thank you! Besides the landscapes, is there anything to see in downtown Reykjavik?"}}

    \vspace{2mm}
    \texttt{[Text response] Yes, besides the landscapes, there are several other attractions to see in Reykjavik, the capital city of Iceland. Some of these include: ...}

    \vspace{5mm}
    \texttt{>\hspace{0mm}>\hspace{0mm}> [Audio Input] \textit{"It's said that there is a very famous blue lagoon near Reykjavik, how much is a ticket?"}}

    \vspace{2mm}
    \texttt{[Text response] The cost of a trip to Blue Lagoon in Iceland can vary depending on several factors including the time of year, accommodation options, and the length of your stay ...}

    \vspace{5mm}
    \texttt{>\hspace{0mm}>\hspace{0mm}> [Audio Input] \textit{"How long does it take from Reykjavik to the glacier lagoon Jokulsarlon by car?"}}

    \vspace{2mm}
    \texttt{[Text response] Sure, here's a short and informative answer to your question, to get from Reykjavik to the Glacier Lagoon Jökulsárlón by car, you can take the following route ...}
\end{tcolorbox}

\end{enumerate}

\end{document}